\documentclass[runningheads]{llncs}
\usepackage[T1]{fontenc}
\usepackage{graphicx}
\usepackage{subcaption}
\usepackage{booktabs}
\usepackage{hyperref}
\usepackage{amsmath}
\usepackage{caption}
\usepackage{changes}
\usepackage{comment}
\usepackage{tikz}
\definecolor{clusterred}{RGB}{254, 192, 193}
\definecolor{clusterorange}{RGB}{254, 217, 193}
\definecolor{clusteryellow}{RGB}{255, 232, 159}
\definecolor{clusterpurple}{RGB}{213, 193, 225}
\definecolor{clusterblue}{RGB}{192, 212, 230}
\definecolor{clustergreen}{RGB}{209, 244, 208}
\newcommand{\colorcircle}[2]{\raisebox{-2pt}{\begin{tikzpicture}[scale=0.2] \node[circle, fill=#1, font=\footnotesize, inner sep=0.35mm]{#2} ; \end{tikzpicture}}}

\begin{document}
\title{In Data or Invisible: Toward a Better Digital Representation of Low-Resource Languages with Knowledge Graphs}
\titlerunning{Toward a Better Representation of Low-Resource Languages with KGs}
%
\author{Ndeye-Emilie Mbengue 
\orcidID{0009-0002-1289-6417} }
\authorrunning{N.-E. Mbengue}

\institute{Université Côte d'Azur, Inria, CNRS, I3S, Sophia-Antipolis, France \\
\email{ndeye-emilie.mbengue@inria.fr}}

\maketitle   
\begin{abstract}
Emerging digital technologies are exacerbating the existing divide in Open Access Data (OAD) between high- and low-resource languages, excluding many communities from participating in the global digital transformation.

In this PhD proposal, we aim to address this gap, focusing on the language coverage of Linked Open Data knowledge graphs (LOD KGs). First, we identify key variables that characterize language distribution in LOD, including the number of Wikipedia articles per language edition and the number of language-tagged entities in LOD KGs. These variables are analyzed across three major multilingual LOD KGs, DBpedia, BabelNet, and Wikidata, providing insights into the representation and distribution of languages within LOD. Building on this analysis, we intend to study the impact of cross-lingual transfer candidate selection on the task of multilingual KG completion. In particular, we plan to investigate strategies based on linguistic proximity and the availability of curated annotated alignments between languages. 
Language proximity also motivates us to explore the benefits of analogical reasoning that relies on (dis)similarities and has not yet been investigated to identify correspondences across languages to improve KG completion performance and enhance language coverage in LOD.
\end{abstract}

\section{Introduction and Motivation}

Digital technologies have significantly transformed contemporary societies, becoming more and more essential in various sectors, including economy~\cite{li_digital_2024}, healthcare~\cite{loo_recommendations_2025}, industry~\cite{li_digital_2024}, and education~\cite{haleem_understanding_2022}. 
As exemplified below, their adaptation to diverse contexts, which reflect the diverse languages, cultures, habits, and behaviors of local populations, has become crucial for countries' development, global visibility, and digital sovereignty.
However, existing socioeconomic factors entail digital access inequalities, excluding some communities from the global digital transformation landscape~\cite{lythreatis_digital_2022}. 

Some digital technologies, and particularly emerging ones such as Generative AI, heavily rely on Open Access Data (OAD)~\cite{wang_open_2025}, \textit{i.e.}, information that is freely shared online for use, modification, and redistribution~\cite{wang_2022}. 
Since a significant divide exists among languages in terms of OAD resources coverage~\cite{Viksna2022AssessingMO,Wolk2004TheEO}, with a high number of low-resource languages, the aforementioned digital access inequalities are likely to be further exacerbated. 
For instance, in 2023, Artificial-Intelligence-driven language technologies remained predominantly limited to the top 3\% of the world’s most widely spoken, politically and economically dominant languages~\cite{helm_diversity_2023}, which often leads Generative AI technologies to suggest and prioritize biased Western-centric content to users, regardless of the language or the cultural context of their prompts~\cite{Naous2023HavingBA}. Addressing the language gap in OAD has thus emerged as a critical, global, and ethical concern toward the preservation of low-resource civilizations and the development of adapted digital technologies for their countries.

Linked Open Data Knowledge Graphs (LOD KGs), defined as graphs of data intended to accumulate and convey knowledge, where nodes represent entities and edges capture diverse relationships between them~\cite{2021Hogan}, offer the opportunity to address the digital gap. Indeed, these graphs provide an abstract and unified representation of multilingual and heterogeneous data modalities that support logical reasoning and cross-lingual transfer to increase their coverage. 
Improving LOD KGs language coverage can thus help in: $(i)$ reducing the gap in OAD by generating new data modalities (\textit{e.g.}, corpora through knowledge graph verbalization~\cite{song-etal-2025-multilingual-verbalisation}). $(ii)$ promoting better digital access and fairness, since they also underpin many digital applications, including established ones such as search engines or recommendation systems~\cite{Peng2023KnowledgeGO} and emerging ones such as GraphRAG in information systems~\cite{graphrag-survey-2025}.

The objective of this PhD is to improve the language coverage in LOD by providing cross-lingual completion methodologies specifically suited to low-resource languages.
Section~\ref{section:sota} reviews the state of the art for: $(i)$ the characterization of languages’ digital coverage, to identify potential low-resource language candidates for the study, and understand the language distribution in LOD. 
$(ii)$ knowledge graph completion methodologies in low-resource settings, and $(iii)$ cross-lingual transfer studies. 
Section~\ref{section:problem-formulation} outlines the problem formulation and the research questions derived from the gaps identified in the state of the art. Finally, Sections~\ref{section:research-methodology-approach},~\ref{section:eval-plan}, and~\ref{section:results} present the planned research methodology, evaluation approach, and expected contributions to address the research questions.

\section{State of the Art}
\label{section:sota}

To better understand how language coverage is tackled within LOD contributions, a systematic literature review was performed using the SciLEx tool~\cite{Ringwald2025ASR} on HAL, Semantic Scholar, Elsevier, and IEEE Explore libraries. This section presents publications from 2020 to 2025 that focus on: $(i)$ the categorization of languages based on their digital coverage, $(ii)$ knowledge graph completion methodologies, and $(iii)$ cross-lingual transfer studies. We used two set of keywords $KW_{1}$= \{``Low-Resource Language'', ``Under-Represented Language'', ``Less-Resourced Language'', ``Cross-lingual'', ``Multilingual''\} and $KW_{2}$= \{``Named Entity Recognition'', ``Relation Extraction'', ``Information Extraction'', ``Transfer'',  ``Knowledge Graph'', ``KG'', ``Completion'', `Enrichment''\}. The search queries were designed such that each document must contain at least one keyword from $KW_1$, while keywords from $KW_2$ were optional. These queries were used to retrieve a list of journal papers and conference articles containing a DOI, keywords, and abstracts. The resulting publications were later manually filtered based on the content of the abstract.\footnote{The final list of reviewed articles is available on \url{https://www.zotero.org/groups/6486731/slr_low_resource_languages_kgs_representation/library}}
While this PhD primarily focuses on LOD, contributions from Natural Language Processing (NLP) are also considered. 
Many LOD KGs are built and enriched through the NLP tasks of Named Entity Recognition (NER) and relation extraction (RE)~\cite{deng_information_2024}, and the language coverage divide is fundamentally tied to the language processing field. 
Incorporating NLP research is therefore essential to better understand language coverage and to leverage NLP techniques for enriching multilingual LOD KGs.

\subsection{Characterization of Language Digital Coverage}
Most reviewed studies in LOD and NLP adopt a dichotomous categorization between high- and low-resource languages, where the decision criteria vary a lot. For instance, Zhong et al.~\cite{zhong_opportunities_2025} survey NLP contributions targeting low-resource languages, providing qualitative characteristics such as ``less studied'', ``resource scarce'', or ``low density'' that are difficult to measure and quantify. 
Similarly, Nigatu et al.~\cite{nigatu_zenos_2024} report qualitative criteria including socio-political characteristics, resource characteristics (\textit{e.g.}, human resources, digital device access), and artifact characteristics (\textit{e.g.}, linguistic knowledge, data, NLP technology) from the analysis of 150 ACL Anthology papers. Only Joshi et al.~\cite{joshi_state_2020} propose a quantitative approach to characterize language digital coverage based on the quantity of annotated corpora available in LDC\footnote{\url{https://catalog.ldc.upenn.edu/}} and ELRA\footnote{\url{https://catalogue.elra.info}}, vs. the quantity of articles available in the different language editions of Wikipedia. 
A fine-grained analysis has not yet been proposed for LOD, which is essential for understanding the language landscape in LOD and selecting the languages to study in this PhD. Extending their work thus represents a first stage in this PhD.

\subsection{Knowledge Graph Completion}

Knowledge Graph Completion (KGC) aims to increase KG coverage by predicting missing entities, types, or relations~\cite{paulheim17}. Recent approaches model KG structural features in a latent space~\cite{aliBHVGSFTL22}, where missing entities and relations can be inferred with scores calculated with, e.g., distances. 
Monolingual approaches perform poorly for languages with scarce KGs~\cite{mao_multilingual_2025}.
Multilingual Knowledge Graph Completion (MKGC) paradigm~\cite{chen_multilingual_2020} improves performance with a learning process that considers jointly multiple KGs in different languages and  manually annotated correspondences (seed alignments). 
MKGC methods can be categorized as: {\it (i)} graph-only, relying solely on topological structure~\cite{tang_multilingual_2023,chen_multilingual_2020}; $(ii)$ text-only, exploiting entity names and descriptions via pretrained language models~\cite{mao_multilingual_2025,song_multilingual_2023}; and $(iii)$ hybrid, combining structural and textual information~\cite{dong_disentangled_2025,huang_multilingual_2022,tong_joint_2022}. Hybrid approaches are preferred in recent contributions as they mitigate limitations of graph-only methods by incorporating multilingual knowledge from language models' large pretraining corpora, especially during the initialization phase.

Some questions remained open in the reviewed contributions. $(i)$ How to ensure the quality of seed alignments? $(ii)$ How to handle only a limited number of seed alignments? 
In this regard, weakly supervised methodologies relying on the distance between seeds in the latent space tend to struggle in resource-scarce scenarios, as they rely heavily on sufficient training data.
Logic-based approaches, particularly those leveraging inductive or abductive reasoning, have shown greater robustness and appear to be more suitable for low-resource language settings~\cite{anil-etal-2024-inductive,AKHTAR2025109660}. 
Still, none of the reviewed studies have explored analogical reasoning despite its reliance on similarity and dissimilarity between objects, which makes it intuitively relevant for extrapolation and transfer, and its demonstrated effectiveness in KG~\cite{zeroshot} and NLP tasks~\cite{alsaidi-2021-analogy,marquer-morpho-2024}.
$(iii)$ How to select the language candidates for joint learning? Most reviewed contributions rely solely on two benchmarks, DBP-5L~\cite{chen_multilingual_2020} and EPKG~\cite{huang_multilingual_2022}, with no evaluation of the choice of the fixed set of transfer language candidates. Addressing these gaps forms a critical second stage of this PhD research.

\subsection{Cross-lingual Transfer}
Cross-lingual transfer is often explored in NLP tasks to tackle low-resource languages. 
We identified three main approaches in the reviewed contributions. $(i)$ Data augmentation, using methods such as back-translation or rule-based techniques~\cite{noauthor_odda_nodate,kolluru_alignment-augmented_2022,liu_improving_2025}, generates additional parallel corpora by leveraging bilingual dictionaries with a high-resource language. $(ii)$ Cross-lingual adaptation applies transfer learning by fine-tuning models trained on high-resource languages~\cite{jin_cycleoie_2025,papadopoulos_penelopie_2021,song-etal-2025-multilingual-verbalisation,soto-martinez-etal-2023-phylogeny}. $(iii)$ Multilingual learning, as for MKGC, leverages models trained across multiple languages to improve low-resource performance~\cite{aharoni-etal-2019-massively,singh_massively_2023,vijayan_language-agnostic_2022}.
Traditional cross-lingual approaches often transfer solely from English, but recent works consider multiple higher-resource pivot languages to reduce hallucinations and cultural bias induced by a single language~\cite{ner_pivot_2025,mohammadshahi_investigating_2024}. Interestingly, the contribution of Martinez et al.~\cite{soto-martinez-etal-2023-phylogeny} showcases the benefits of selecting cross-lingual transfer language candidates based on a language proximity (\textit{e.g.}, phylogenetic tree) in KG verbalization performance.
But no study has yet benchmarked the contribution of several prior proximities to guide the selection of cross-lingual transfer candidates. This constitutes the third stage of this PhD proposal.

\section{Problem Formulation and Research Questions}
\label{section:problem-formulation}

The literature review allowed us to identify several research gaps toward improving language coverage in LOD KGs. First, no fine-grained characterization of language digital coverage in LOD is currently available, although such an analysis is essential to identify critically under-resourced languages, distinguish higher-resource languages, and better understand distribution patterns within LOD. 
Second, MKGC represents a promising direction for enriching low-resource language KGs. However, existing methodologies do not sufficiently investigate the impact of language selection as a preliminary step, nor the effect of choosing appropriate seed alignments. 
Third, analogical reasoning has not yet been proposed as a framework for constructing weakly supervised alignments in scenarios where available seed data is extremely scarce.
Based on these findings, we formulate one main research question and three sub-research questions:
\begin{description} 
    \item[(RQ)]Which criteria and methods are effective at selecting and addressing low-covered languages in LOD?
    \begin{description}
        \item[(SRQ1)] Which variable (or combination) better represents the global language distribution in LOD among the number of available corpora, the number of language-tagged entities, and the number of language-tagged relations in LOD KGs?
        \item[(SRQ2)] Which criterion (or combination) for candidate transfer selection achieves the best trade-off between performance and computational cost in MKGC: the language coverage in LOD, the language proximity, or the volume of available annotated alignments?
        \item[(SRQ3)] Does analogical reasoning provide an effective weakly supervised framework for cross-lingual alignment and completion? What is the impact of the amount of annotated data?
    \end{description}
\end{description}

\section{Research Methodology and Approach}
\label{section:research-methodology-approach}

\subsection{(SRQ1) Representing the Global Language Distribution in LOD}
Following the methodology of Joshi et al.~\cite{joshi_state_2020}, and to facilitate a comparative analysis between NLP and LOD ecosystem, the set of languages considered in this study is restricted to the set of written languages documented in The World Atlas of Language Structure (WALS).\footnote{\url{https://wals.info/languoid}\label{footnote:wals}} Quantitative variables inspired by prior work in NLP, particularly in \cite{joshi_state_2020}, are selected and measured to capture three complementary dimensions of language coverage: $(i)$ the number of articles in the language-specific Wikipedia edition, often used to build or enrich KGs, $(ii)$ the number of entities (or synsets) annotated with the target language in KGs, and $(iii)$ the number of relations annotated with the target language in KGs.

\subsection{(SRQ2) Candidate Transfer Selection}


To assess the impact of language transfer candidate selection and the number of available seed alignments, both in terms of computational cost and model performance, we intend to benchmark several state-of-the-art (SoTA) models on the task of MKGC. 
Those models are reported in Table~\ref{tab:sota_comparison}, which represents their performance on the traditional DBP-5L benchmark, where all the languages of the set are used as transfer candidates. Future experiments will follow three configurations, given a target language and LOD KG:
$(i)$ \textbf{Digital Resource.} Based on the analysis for (SRQ1), transfer candidates are selected from languages with the highest coverage.
$(ii)$ \textbf{Language Proximity.} Transfer candidates are selected based on \textit{a priori} language proximity, such as the language family as in~\cite{soto-martinez-etal-2023-phylogeny}, but also macro-area, genus, and shared linguistic features that are taken from WALS. 
$(iii)$ \textbf{Curated Seed Alignments.} Transfer candidates are selected based on the availability of curated seed alignments between considered LOD KGs. Inspired by the contribution of Luo et al.~\cite{Luo2021GraphEG}, we envision a curation process that would rely on the exclusion of seed alignments that increase the noise of the joined multilingual KGs. The noise could be measured using a minimal entropy approach derived from information theory. Further details and a precise formalization are yet to be investigated.
\begin{table}
    \centering
    \small
    \setlength{\tabcolsep}{6pt}
    \renewcommand{\arraystretch}{0.8}
    \caption{Different State-of-the-Art models and their performance (Hits@1) on DBP-5L traditional transfer groups. A higher value shows better performance.}
    \resizebox{\textwidth}{!}{
    \begin{tabular}{c c c c c c c}
        \toprule
        \textbf{Model} & \textbf{Family} & \textbf{Greek} & \textbf{Japanese} & \textbf{English} & \textbf{Spanish} & \textbf{French}  \\
        \midrule
        KENS~\cite{chen_multilingual_2020}& GRAPH & 27.5 & 32.9 & 14.4&22.3 &25.2 \\
        SS-AGA~\cite{huang_multilingual_2022}& GRAPH & 30.8 & 34.6 & 16.3& 25.5 & 27.1 \\
        \midrule
        AlignKGC~\cite{chakrabarti_joint_2022}& HYBRID  & 57.6 & 53.2 & 37.2&53.0 &52.9 \\
        JMAC~\cite{tong_joint_2022}& HYBRID  & 55.2 & 53.3 & 29.5 & 45.4 & 49.3 \\
        DMGNNSI~\cite{dong_disentangled_2025}& HYBRID  & 63.6&59.7& 38.7 & 57.3 & 60.2 \\
        CA-MKGC~\cite{zhang_conflict-aware_2023}& HYBRID & 59.6 & 54.6 &34.9 &49.1 & 51.3 \\
        \bottomrule
    \end{tabular}}
    \label{tab:sota_comparison}
\end{table}

In the first step, all analyses will be conducted using the languages in DBP-5L as target languages. Both the initial transfer candidates and those generated from the experiments will be considered in the evaluation.
In the second step, experiments will focus on target languages identified as low-resource, based on the analysis of (SRQ1). This dual evaluation aims at highlighting the variables that enhance cross-lingual transfer performance in MKGC and to determine which models are most effective in low-resource language settings. Our experiments will provide new datasets for MKGC, and the insights obtained, especially from the curated seed alignments analysis, will be leveraged to improve the alignment generation in a weakly supervised paradigm (SRQ3).

\subsection{SRQ3) Analogical Reasoning for Cross-Lingual Alignment}

Analogical reasoning, based on proportions of the form ``A is to B as C is to D'' (\textit{e.g.}, ``Star'' is to ``Stars" as ``Moon'' is to ``Moons''), enables humans to efficiently learn cross-lingual correspondences. 
This reasoning can thus be leveraged to build a weakly supervised alignment framework that guides the task of MKGC by inferring additional alignments, or at least close correspondences, between target and transfer languages, leveraging existing alignments between them as well as alignments across other languages with a higher volume of curated aligned pairs. The goal is to learn a language-agnostic model capable of detecting multilingual correspondences. 
As a first baseline, we will rely on the analogical classifier proposed by Lim et al.~\cite{Lim2019SolvingWA}, which employs a two-layer CNN, and will extend it in future work.

\section{Evaluation Plan}
\label{section:eval-plan}
Three early-stage research methodologies have been developed to address the different sub-research questions. The first methodology identifies the variables to be evaluated for the characterization of language digital coverage. The second defines how language groups and curated seed alignments are likely to be selected, as well as the state-of-the-art models to be used for MKGC. Finally, the third methodology presents the initial architecture for an analogy classifier designed to detect multilingual correspondences. The following section describes the methodologies that will be used to assess these contributions.

\subsection{Representing the Global Language Distribution in LOD}

To assess which variable among $(i)$ the number of articles in the language-specific Wikipedia edition, $(ii)$ the number of language-tagged relations, $(iii)$ the number
of relations annotated with the target language in LOD KGs, or $(iv)$ their combination, best captures the global language distribution in LOD, we adopt a multi-step evaluation strategy. 
First, we apply an unsupervised clustering algorithm to group languages into categories based on their values for the previous variables.
Next, we evaluate the quality of these different categorizations using traditional clustering validation metrics. In particular, we compute the Average Silhouette Score (ASS) to measure the cohesion and separation of language groups, and the Variance Ratio Index (VRI) to assess inter-cluster dispersion relative to intra-cluster variance~\cite{HAN2012443}. 
Additionally, we compare the resulting categorizations with existing language classification frameworks, such as the taxonomy proposed by Joshi et al.~\cite{joshi_state_2020}. This alignment is evaluated using the Adjusted Rand Index (ARI) and Normalized Mutual Information (NMI).
As a result of the evaluation, we aim to retain the most effective combination of variables to propose a categorization of languages in LOD. To ensure consistency and reproducibility across different data sources and domains, a quantile-based approach is preferred to define categories and their boundaries, as it systematizes the methodology and facilitates the extension of the analysis to additional data sources and domains.

\subsection{Multilingual Knowledge Graph Completion}

To evaluate and compare the effectiveness of the different experiments on the MKGC task, we first adopt standard evaluation metrics commonly used in link prediction. In particular, we will report the Mean Reciprocal Rank (MRR), which captures the average inverse rank of the correct entity, and Hits@$K$, which measures the proportion of correct entities ranked within the top-$K$ predictions. These metrics provide complementary insights into both overall ranking quality and top-$K$ accuracy.
In addition to predictive performance, we assess computational efficiency by measuring training time and inference time. These metrics allow us to evaluate the scalability of the approach and quantify the trade-off between improved multilingual representation and computational cost.

\section{Preliminary Results}
\label{section:results}
\begin{figure*}[t]
    \centering

    \begin{subfigure}[b]{0.45\textwidth}
        \centering
        \includegraphics[width=\linewidth]{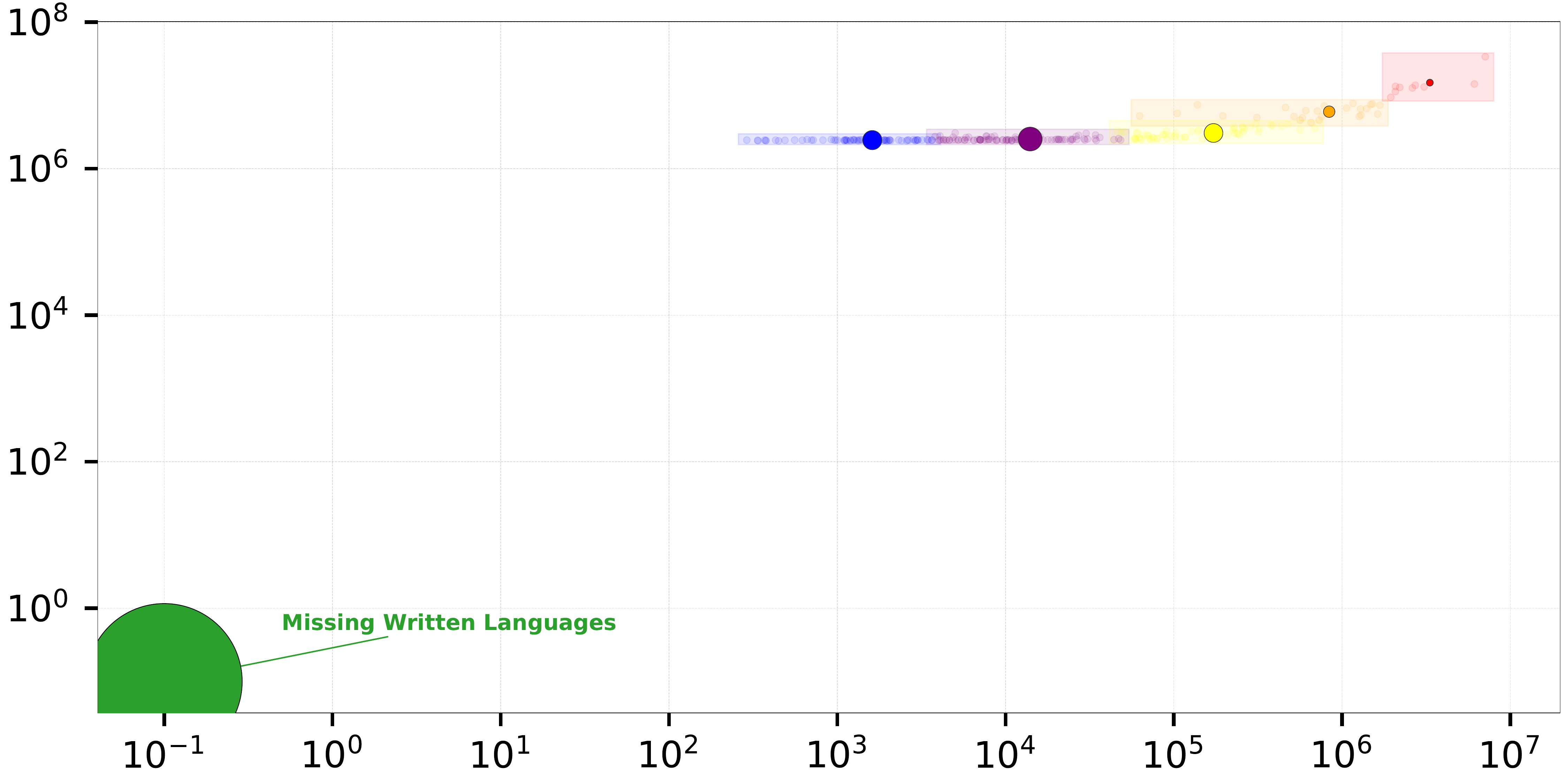}
        \caption{BabelNet-based categories.}
        \label{fig:babelnet}
    \end{subfigure}
    \hfill
    \begin{subfigure}[b]{0.45\textwidth}
        \centering 
        \includegraphics[width=\linewidth]{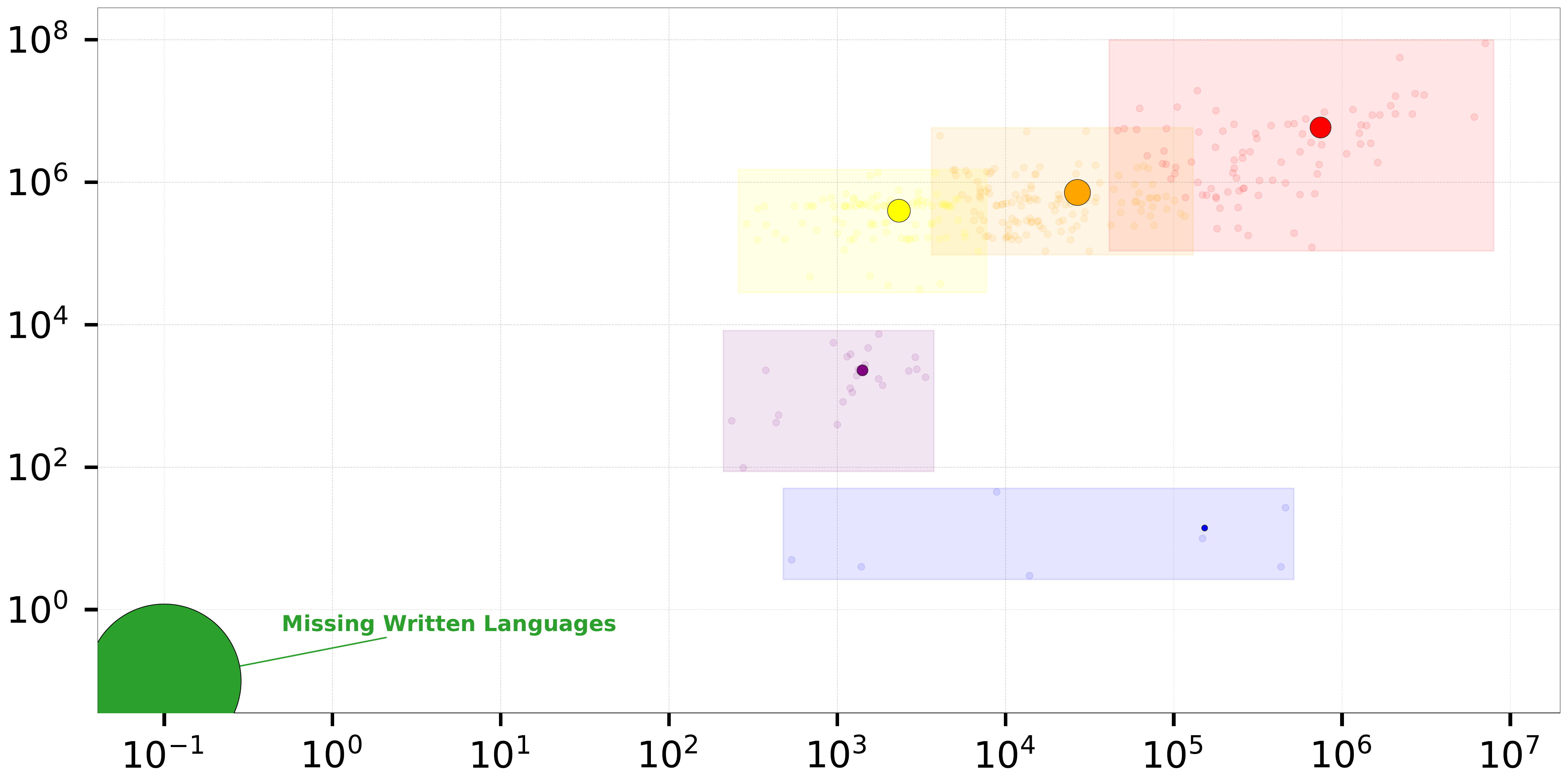}
        \caption{Wikidata-based categories.}
        \label{fig:wikidata}
    \end{subfigure}

    \vspace{0.3cm}

    \begin{subfigure}[b]{0.45\textwidth}
        \centering
        \includegraphics[width=\linewidth]{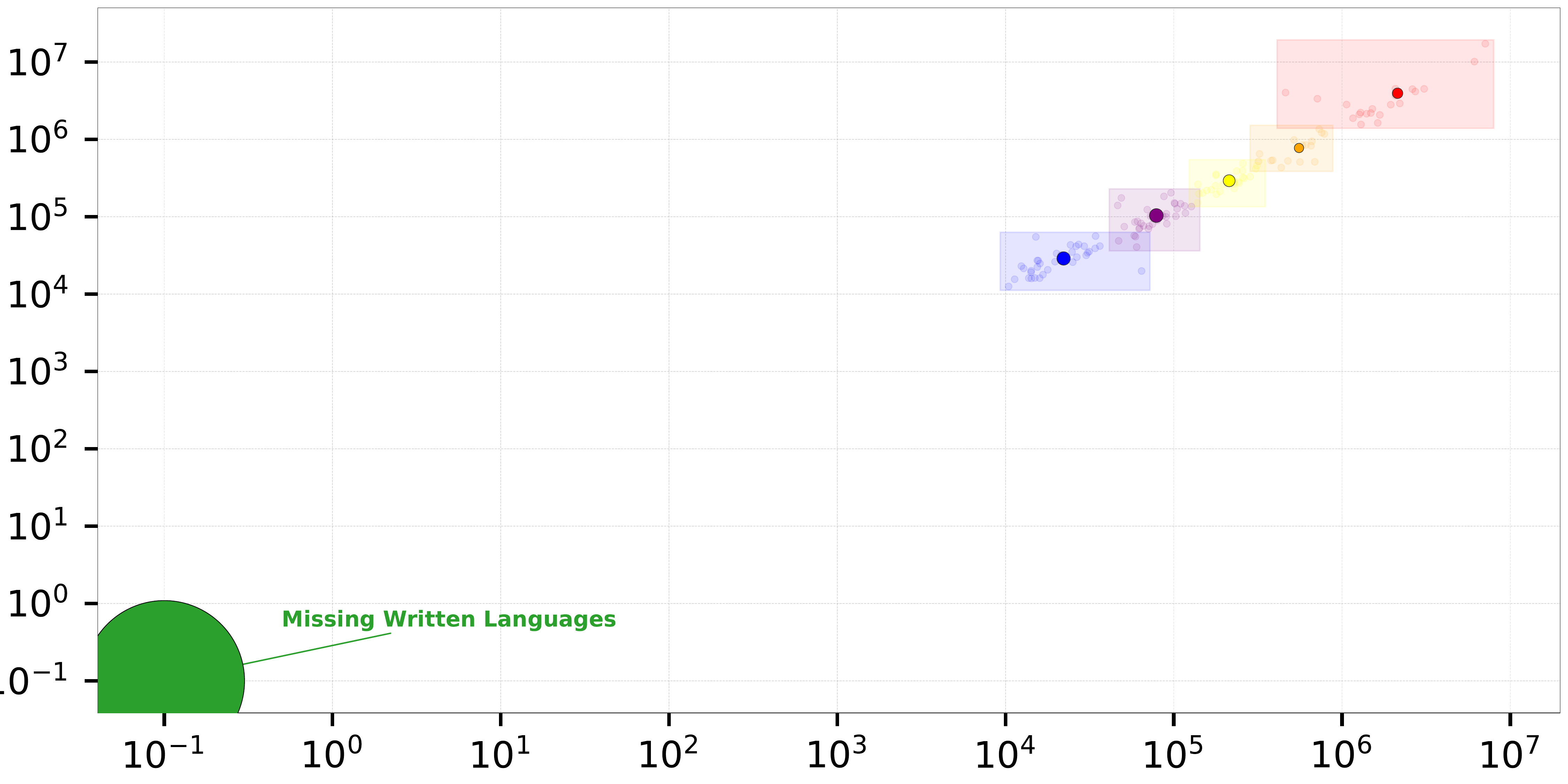}
        \caption{DBpedia-based categories.}
        \label{fig:dbpedia}
    \end{subfigure}
    \hfill
    \begin{subfigure}[b]{0.45\textwidth}
        \centering
        \includegraphics[width=\linewidth]{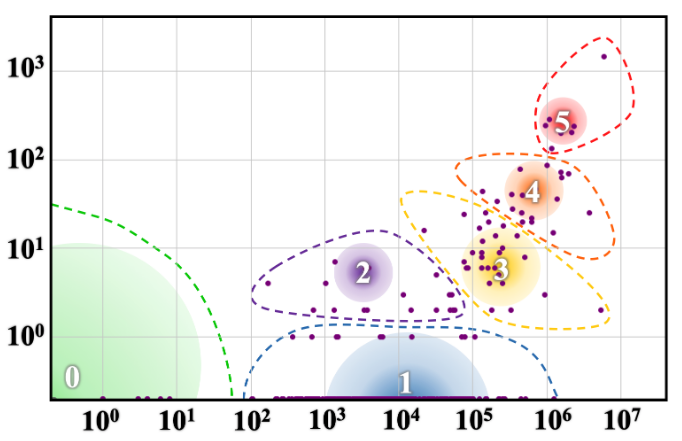}
        \caption{Joshi et al. categories.}
        \label{fig:joshicat}
    \end{subfigure}

    \caption[]{Language coverage (log-log scale) in BabelNet, Wikidata, and DBpedia, compared with the coverage in NLP corpora obtained by Joshi et al.~\cite{joshi_state_2020}. The x-axis represents the number of articles in the Wikipedia language edition, while the y-axis represents the count of language-tagged entities in KGs, or the number of available annotated corpora for this language. 
    Clusters and colors from~\cite{joshi_state_2020}: \colorcircle{clusterred}{5}~Winner, \colorcircle{clusterorange}{4}~Underdogs, \colorcircle{clusteryellow}{3}~Rising Stars, \colorcircle{clusterpurple}{2}~Hopefuls, \colorcircle{clusterblue}{1}~Scrapping-Bys, \colorcircle{clustergreen}{0}~Left-Behinds.
    }
    \label{fig:clusters_comparison}
\end{figure*}

The evaluation section outlines different approaches to address the three sub-research questions, highlighting the variables, criteria, and algorithms that support the analysis and enhancement of language coverage in LOD KGs. As this PhD is still in its early stages, the only contribution so far focuses on the analysis of language distribution in LOD, based on the variables established in the methodology section for (SRQ1).
To provide an initial overview of language distribution in LOD, we leverage $(i)$ the number of articles in a language-specific Wikipedia edition and $(ii)$ the number of entities or synsets annotated in the target language within three LOD KGs: BabelNet, Wikidata, and DBpedia. 
To compare this distribution in LOD KGs vs. the one in NLP from Joshi et al.~\cite{joshi_state_2020}, we combine these two variables and use an unsupervised clustering methodology (K-means) to build six categories of languages, displayed in Figure~\ref{fig:clusters_comparison}.
A website\footnote{\label{footnote:webpage}\url{https://nembengue.github.io/language_digital_coverage_lod/ }} provides dynamic versions of these visualizations and additional ones.
Across all three KGs, the distribution of languages is skewed: a small number of languages dominate both Wikipedia and LOD KGs, while the vast majority have minimal or no representation. 
The size of the artificially-induced green cluster, representing written languages from WALS that lack coverage in both KGs and Wikipedia, quantitatively highlights the exclusion of a majority of languages and communities of the World from the Linked Open Data.
This is aligned with the findings observed by Joshi et al.~\cite{joshi_state_2020} in the NLP community.

The three LOD KGs exhibit distinct distribution patterns. 
DBpedia follows a near-linear, stratified trend, reflecting its reliance on Wikipedia infoboxes extraction. This linearity indicates that coverage is tightly linked with Wikipedia resources and activities. 
BabelNet shows a more homogeneous distribution with less dispersion. 
Its integration from both Wikipedia and WordNet could explain the non-linear pattern observed compared to DBpedia.
Note that the skew remains, as a huge amount of languages are still not covered.
As for Wikidata, this KG exhibits a highly heterogeneous and imbalanced landscape, with clusters of varying size. 
This distribution highlights that, as for Wikipedia, Wikidata coverage depends on the editorial community's devotion.
Based on those variables, the language distribution in LOD is very heterogeneous and different from the one observed in NLP. To illustrate, the normalized mutual information scores computed using Joshi et
al.’s categorization as reference is 0.63 for DBpedia, 0.60 for BabelNet, and 0.56
for Wikidata.

The different deviations from the linear trend also confirm some insights.
Right-divergent clusters correspond to languages with strong Wikipedia coverage but weak LOD KG coverage, suggesting a potential for enrichment via information extraction from text corpora.
Left-divergent clusters correspond to languages with strong KG presence but limited textual resources, highlighting the opportunity for KG verbalization to generate additional corpora.
Clusters aligned with the linear trend have a similar amount of Wikipedia and LOD coverage and are likely to benefit most from LOD KGs cross-lingual transfer.

The language distribution also reveals additional opportunities for knowledge transfer. In particular, our results show that language families are homogeneously distributed across clusters (available online\textsuperscript{\ref{footnote:webpage}}). This suggests that lower resource languages can benefit from transfer from higher resource languages within the same family. Such family-based transfer groups can support knowledge transfer and improve performance in MKGC (SRQ2).
For instance, based on the aggregation of entity counts across the three LOD KGs, candidate low-resource languages that could benefit from this approach include Neapolitan and Ladino, which belong to the Romance language genus.

Next, these language categories will be compared with those obtained using other (combinations of) variables, such as the number of language-tagged relations in LOD KGs, to evaluate them in terms of cohesion and consistency. This comparison will provide insights into the most informative variables for characterizing languages and will serve as a foundation for developing approaches for MKGC. A formalization of the different levels of resource availability will also be developed to establish systematized definitions of low-resource languages in LOD.

\section{Conclusion}

The digital language gap, amplified by emerging tools such as Generative AI, is a growing concern as it isolates certain communities and countries from the global digital transformation, which plays a critical role in sectors such as healthcare and education. 
This PhD addresses the digital language gap in Linked Open Data by focusing on low-resource languages and exploring methods to analyze and enrich their coverage. The main research question investigates which criteria and methods are effective at identifying and addressing low-resource languages in LOD. Based on the gaps observed in the literature review, three sub-research questions have been established. (SRQ1) focuses on identifying which variable or combination of variables best represents global language distribution in LOD. Using preliminary results based on Wikipedia article counts and language-tagged entities in DBpedia, Wikidata, and BabelNet, we observe a highly skewed distribution heterogeneous within LOD sources and a previous NLP distribution analysis. These initial findings will be assessed and evaluated using other variables, such as language-tagged relations, to determine the most informative and reliable metrics. The potential outcome of (SRQ1) is to provide the foundation for a generalizable and formal definition of low-resource languages in LOD, guiding subsequent enrichment and transfer strategies. (SRQ2) explores how to enrich low-resource language knowledge graphs by selecting appropriate transfer candidates. The selection can be guided by language coverage in LOD, linguistic proximity, and the volume of curated seed alignments. Benchmarking different configurations will highlight which criterion or combination of criteria maximizes performance while controlling computational cost in multilingual knowledge graph completion. (SRQ3) investigates whether analogical reasoning can provide a weakly supervised framework to infer cross-linguistic correspondences and support MKGC in low-resource scenarios.

Overall, this PhD, rooted in Linked Open Data standards, will provide a framework for analyzing, enriching, and generating resources for low-resource languages to strengthen a more equitable digital access.

\begin{credits}
\subsubsection{\ackname}

I would like to thank my PhD supervisors, Fabien Gandon, Pierre Monnin, and Miguel Couceiro, for their guidance and patience through the first stages of this PhD. Their methodological advice and meticulous feedback have helped a lot in improving the overall quality of this document.

\end{credits}
%
%
%
%
\bibliography{sources}
\end{document}